\documentclass[10pt,twocolumn,letterpaper]{article}

\usepackage[pagenumbers]{cvpr} 

\usepackage[accsupp]{axessibility}
\usepackage{graphicx}
\usepackage{amsmath}
\usepackage{amssymb}
\usepackage{booktabs}
\usepackage{amsthm}
\usepackage{subcaption}
\usepackage{bm}
\usepackage{cases}
\usepackage{threeparttable}
\usepackage{diagbox}
\usepackage{multirow}

\newtheorem{proposition}{Proposition}
\newcommand{\thickhline}{%
	\noalign {\ifnum 0=`}\fi \hrule height 1pt
	\futurelet \reserved@a \@xhline
}

\usepackage[pagebackref,breaklinks,colorlinks]{hyperref}

\usepackage{algorithm}
\usepackage{algorithmicx}
\usepackage{algpseudocode}
\floatname{algorithm}{Algorithm}

\usepackage[capitalize]{cleveref}
\crefname{section}{Sec.}{Secs.}
\Crefname{section}{Section}{Sections}
\Crefname{table}{Table}{Tables}
\crefname{table}{Tab.}{Tabs.}

\begin{document}

\title{Training High-Performance Low-Latency Spiking Neural Networks by Differentiation on Spike Representation}

\author{
	Qingyan Meng$^{1,2}$, Mingqing Xiao$^3$, Shen Yan$^{4}$, Yisen Wang$^{3,5}$, Zhouchen Lin$^{3,5,6,}$\thanks{Corresponding author.} , Zhi-Quan Luo$^{1,2}$\\
	$^1$The Chinese University of Hong Kong, Shenzhen \
	$^2$Shenzhen Research Institute of Big Data\\
	$^3$Key Lab. of Machine Perception (MoE), School of Artificial Intelligence, Peking University \\
	$^4$Center for Data Science, Peking University \
	$^5$Institute for Artificial Intelligence, Peking University\\
	$^6$Peng Cheng Laboratory\\
	{\tt \small qingyanmeng@link.cuhk.edu.cn, \{mingqing\_xiao, yanshen, yisen.wang, zlin\}@pku.edu.cn,} \\ {\tt \small luozq@cuhk.edu.cn}
}

\maketitle

\begin{abstract}
	\vspace{-7pt}
   Spiking Neural Network (SNN) is a promising energy-efficient AI model when implemented on neuromorphic hardware. However, it is a challenge to efficiently train SNNs due to their non-differentiability. Most existing methods either suffer from high latency (\emph{\ie}, long simulation time steps), or cannot achieve as high performance as Artificial Neural Networks (ANNs). In this paper, we propose the Differentiation on Spike Representation (DSR) method, which could achieve high performance that is competitive to ANNs yet with low latency. First, we encode the spike trains into spike representation using (weighted) firing rate coding. Based on the spike representation, we systematically derive that the spiking dynamics with common neural models can be represented as some sub-differentiable mapping. With this viewpoint, our proposed DSR method trains SNNs through gradients of the mapping and avoids the common non-differentiability problem in SNN training. Then we analyze the error when representing the specific mapping with the forward computation of the SNN. To reduce such error, we propose to train the spike threshold in each layer, and to introduce a new hyperparameter for the neural models. With these components, the DSR method can achieve state-of-the-art SNN performance with low latency on both static and neuromorphic datasets, including CIFAR-10, CIFAR-100, ImageNet, and DVS-CIFAR10. 
\end{abstract}

\vspace{-8pt}
\section{Introduction}
\label{sec:intro}
\vspace{-4pt}
Inspired by biological neurons that communicate using spikes, Spiking Neural Networks (SNNs) have recently received surging attention. This promise depends on their energy efficiency on neuromorphic hardware~\cite{merolla2014million,davies2018loihi,pei2019towards}, while deep Artificial Neural Networks (ANNs) require substantial power consumption.

However, the training of SNNs is a major challenge \cite{tavanaei2019deep} since information in SNNs is transmitted through non-differentiable spike trains. Specifically, the non-differentiability in SNN computation hampers the effective usage of gradient-based backpropagation methods. To tackle this problem, the surrogate gradient (SG) method \cite{neftci2019surrogate,wu2018spatio,shrestha2018slayer,fang2021sew,zheng2020going} and the ANN-to-SNN conversion method \cite{cao2015spiking,deng2021optimal,sengupta2019going,rueckauer2017conversion,yan2021near} have been proposed and yielded the best performance. In the SG method, an SNN is regarded as a recurrent neural network (RNN) and trained by the backpropagation through time (BPTT) framework. And during backpropagation, gradients of non-differentiable spike functions are approximated by some surrogate gradients. Although the SG method can train SNNs with low latency (\ie, short simulation time steps), it cannot achieve high performance comparable to leading ANNs. Besides, the adopted BPTT framework needs to backpropagate gradients through both the layer-by-layer spatial domain and the temporal domain, leading to a long training time and high memory cost of the SG method. The high training costs further limit the usage of large-scale network architectures. On the other hand, the ANN-to-SNN conversion method directly determines the network weights of an SNN from a corresponding ANN, relying on the connection 
between firing rates of the SNN and activations of the ANN. The conversion method enables the obtained SNN to perform as competent as its ANN counterpart. However, intolerably high latency is typically required, since only a large number of time steps can make the firing rates closely approach the high-precision activation values of ANNs \cite{rueckauer2017conversion,hu2018spiking}. Overall, SNNs obtained by the two widely-used methods either cannot compete their ANN counterparts, or suffer from high latency.

\begin{table}[]
	\caption{Comparison of the ANN-to-SNN conversion, surrogate gradient (SG), and DSR method with respect to latency, performance with low latency, and applicability on neuromorphic data.}
	\label{table:method-comparison}
	\centering
	\vspace{-6pt}
	\begin{tabular}{cccc}
		\hline  & Conversion & SG & DSR \\
		\hline 
		Latency & High & \textbf{Low} & \textbf{Low} \vspace{2pt}  \\
		Performace w/ & \multirow{2}*{Low} & \multirow{2}*{Medium} & \multirow{2}*{\textbf{High}} \vspace{-2pt}\\
		Low Latency & & & \vspace{2pt} \\
		Neuromorphic & Non-appli- & \textbf{Appli-} & \textbf{Appli-} \vspace{-2pt} \\
		Data& cable& \textbf{cable} & \textbf{cable} \\
		\hline
	\end{tabular}
	\vspace{-8pt}
\end{table}

In this paper, we overcome both the low performance and high latency issues by introducing the Differentiation on Spike Representation (DSR) method to train SNNs. First, we treat the (weighted) firing rate of the spiking neurons as spike representation. Based on the representation, we show that the forward computation of an SNN with common spiking neurons can be represented as some sub-differentiable mapping. We then derive the backpropagation algorithm while treating the spike representation as the information carrier. In this way, our method encodes the temporal information into spike representation and backpropagates through sub-differentiable mappings of it, avoiding calculating gradients at each time step. To effectively train SNNs with low latency, we further study the representation error due to the SNN-to-mapping approximation, and propose to train the spike thresholds and introduce a new hyperparameter for the spiking neural models to reduce the error. With these methods, we can train high-performance low-latency SNNs. And the comparison of the properties between the DSR method and other methods is illustrated in \cref{table:method-comparison}. Formally, our main contributions are summarized as follows:
\begin{itemize}
	\vspace{-3.5pt}
	\item[1.] We systematically study the spike representation for common spiking neural models, and propose the DSR method that uses the representation to train SNNs by backpropagation.
	The proposed method avoids the non-differentiability problem in SNN training and does not require the costly error backpropagation through the temporal domain.
	\vspace{-4pt}
	\item[2.] We propose to train the spike thresholds and introduce a new hyperparameter for the spiking neural models to reduce the representation error. The two techniques greatly help the DSR method to train SNNs with high performance and low latency. 
	\vspace{-4pt}
	\item[3.] Our model achieves competitive or state-of-the-art (SOTA) SNN performance with low latency on the CIFAR-10\cite{krizhevsky2009learning}, CIFAR-100\cite{krizhevsky2009learning}, ImageNet\cite{deng2009imagenet}, and DVS-CIFAR10\cite{li2017cifar10} datasets. Furthermore, the experiments also prove the effectiveness of the DSR method under ultra-low latency or deep network structures.
\end{itemize}

\vspace{-10pt}
\section{Related Work}
\label{sec:related}
\vspace{-3pt}

Many works seek biological plausibility in training SNNs \cite{caporale2008spike,kheradpisheh2018stdp,legenstein2008learning} using derivations of the Hebbian learning rule \cite{hebb1949organisation}. However, this method cannot achieve competitive performance and cannot be applicable on complicated datasets. Besides the brain-inspired method, SNN learning methods can be mainly categorized into two classes:  ANN-to-SNN conversion \cite{deng2021optimal,yan2021near,sengupta2019going,rueckauer2017conversion,hunsberger2015spiking,diehl2015fast,kim2020spiking,han2020rmp,han2020deep,ding2021optimal} and direct training\cite{bohte2000spikeprop,huh2017gradient,zhang2020temporal,zheng2020going,mostafa2017supervised,wu2019direct,wu2018spatio,bellec2018long,neftci2019surrogate,yang2021backpropagated,xiao2021ide,shrestha2018slayer,fang2021sew,esser2015backpropagation,esser2016convolutional}. We discuss both the conversion and direct training method, then analyze the information representation used in them. 

\vspace{-12pt}
\paragraph{ANN-to-SNN Conversion} The feasibility of this conversion method relies on the fact that the firing rates of an SNN can be estimated by activations of an ANN with corresponding architecture and weights\cite{rueckauer2017conversion}. With this method, the parameters of a target SNN are directly determined from a source ANN. And the performance of the target SNN is supposed to be not much worse than the source ANN. Many effective techniques have been proposed to reduce the performance gap, such as weight normalization\cite{sengupta2019going}, temporal switch coding\cite{han2020deep}, rate norm layer\cite{ding2021optimal}, and bias shift\cite{deng2021optimal}. Recently, the conversion method has achieved high-performance ANN-to-SNN conversion \cite{li2021free,yan2021near,deng2021optimal}, even on ImageNet. However, the good performance is at the expense of high latency, since only high latency can make the firing rates closely approach the high-precision activation. This fact hurts the energy efficiency of SNNs when using the conversion method. Furthermore, the conversion method is not suitable for neuromorphic data. In this paper, we borrow the idea of ANN-SNN mapping to design the backpropagation algorithm for training SNNs. However, unlike usual ANN-to-SNN conversion methods, the proposed DSR method can obtain high performance with low latency on both static and neuromorphic data.

\vspace{-12pt}
\paragraph{Direct Training} Inspired by the immense success of gradient descent-based algorithms for training ANN, some works regard an SNN as an RNN and directly train it with the BPTT method. This scheme typically leverages surrogate gradient to deal with the discontinuous spike functions \cite{bohte2000spikeprop,wu2018spatio,zheng2020going,neftci2019surrogate}, or calculate the gradients of loss with respect to spike times\cite{mostafa2017supervised,wunderlich2020eventprop,zhou2019temporal}. Between them, the surrogate gradient method achieves better performance with lower latency\cite{fang2021sew,zheng2020going}.
However, those approaches need to backpropagate error signals through time steps and thus suffer from high computational costs during training \cite{deng2020rethinking}. Furthermore, the inaccurate approximations for computing the gradients or the ``dead neuron'' problem\cite{shrestha2018slayer} limit the training effect and the use of large-scale network architectures. The proposed method uses spike representation to calculate the gradients of loss and need not backpropagate error through time steps. Therefore, the proposed method avoids the common problems for direct training. A few works \cite{wu2021training,wu2021tandem} also use the similar idea of decoupling the forward and backward passes to train feedforward SNNs;
however, they neither systematically analyze the representation schemes nor the representation error, and they cannot achieve comparable accuracy as ours, even with high latency.

\vspace{-12pt}
\paragraph{Information Representation in SNNs} In SNNs, information is carried by some representation of spike trains \cite{panzeri2001unified}. There are mainly two representation schemes: temporal coding and rate coding. These two schemes treat exact firing times and firing rates, respectively, as the information carrier. Temporal coding is adopted by some direct training methods that calculate gradients with respect to spike times \cite{mostafa2017supervised,wunderlich2020eventprop,zhou2019temporal}, or few ANN-to-SNN methods \cite{han2020deep,stockl2021optimized}. With temporal coding, those methods typically enjoy low energy consumption on neuromorphic chips due to sparse spikes. However, those methods either require chip-unfriendly neuron settings \cite{han2020deep,stockl2021optimized,zhou2019temporal}, or only perform well on simple datasets. Rate coding is adopted by most ANN-to-SNN methods \cite{deng2021optimal,yan2021near,sengupta2019going,rueckauer2017conversion,diehl2015fast,kim2020spiking,han2020rmp,ding2021optimal} and many direct training methods \cite{xiao2021ide,wu2021tandem}. The rate coding-based methods typically achieve better performances than those with temporal coding. Furthermore, recent progress shows the potential of rate-coding based methods on low latency or sparse firing \cite{xiao2021ide}, making it possible to reach the same or even better level of energy efficiency as temporal coding scheme. In this paper, we adopt the rate coding scheme to train SNNs.


\section{Proposed Differentiation on Spike Representation (DSR) Method}
\label{sec:method}

\subsection{Spiking Neural Models}
\label{sec:neuralmodel}
\vspace{-1pt}
Spiking neurons imitate the biological neurons that communicate with each other by spike trains. In this paper, we adopt the widely used integrate-and-fire (IF) model and leaky-integrate-and-fire (LIF) model \cite{burkitt2006review}, both of which are simplified models to characterize the process of spike generation. Each IF neuron or LIF neuron `integrates' the received spike as its membrane potential $V(t)$, and the dynamics of membrane potential can be formally depicted as
\vspace{-12pt}
\begin{align}
&\text{IF:} \ \ \ \ \frac{\mathrm{d} V(t)}{\mathrm{d} t}=I(t), \quad\quad\quad\quad\quad\quad\quad\quad\quad \ V<V_{th}, \label{IF}\\
&\text{LIF:} \ \ \tau \frac{\mathrm{d} V(t)}{\mathrm{d} t}=-(V(t)-V_{rest}) + I(t), \ \ V<V_{th}, \label{LIF}
\end{align}
\vspace{-1pt}
where $V_{rest}$ is the resting potential,  $\tau$ is the time constant, $V_{th}$ is the spike threshold, and $I$ is the input current which is related to received spikes. Once the membrane potential $V$ exceeds the predefined threshold $V_{th}$ at time $t_f$, the neuron will fire a spike and reset its membrane potential to the resting potential $V_{rest}$. The output spike train can be expressed using the Dirac delta function $s(t)=\sum_{t_f}\delta(t-t_f)$.

In practice, discretization for the dynamics is required. The discretized model is described as:
\begin{subnumcases}{} \label{discrete}
U[n]=f(V[n-1],I[n]), \label{discretea} \\
{s}[n]=H(U[n]-V_{th}), \label{discreteb}  \\
V[n]=U[n]-V_{th} {s}[n], \label{discretec}
\end{subnumcases}
where $U[n]$ is the membrane potential before resetting, ${s}[n] \in \{0,1\}$ is the output spike, $n=1,2,\cdots,N$ is the time step index and $N$ is the latency, $H(x)$ is the Heaviside step function, and $f$ is the membrane potential update function. In the discretization, both $V[0]$ and $V_{rest}$ are set to be $0$ for simplicity, and therefore $V_{th}>0$. The function $f(\cdot,\cdot)$ for IF and LIF models can be described as:
\vspace{-3.5pt}
\begin{align}
&\text{IF:} \ f(V,I)=V+I, \label{IF-update} \\
&\text{LIF:} \ f(V,I)=e^{-\frac{\Delta t}{\tau}} V+\left(1-e^{-\frac{\Delta t}{\tau}}\right) I, \label{LIF-update} 
\end{align}
where $\Delta t<\tau$ is the discrete step for LIF model. In practice, we set $\Delta t$ to be much less that $\tau$. Different from other literature \cite{wu2018spatio,fang2021incorporating,deng2021optimal},
we explicitly introduce the hyperparameter $\Delta t$ to ensure a large feasible region for $\tau$, since the discretization for LIF model is only valid when the discrete step $ \tau > \Delta t$ \cite{fang2021incorporating}. For example, $\tau = 1$ is allowed in our setting, while some other works prohibit it since they set $\Delta t = 1$. We use the ``reduce by subtraction'' method \cite{xiao2021ide,wu2021tandem} for resetting the membrane potential in Eq. \eqref{discretec}. Combing Eqs. \eqref{discretea} and \eqref{discretec}, we get a more concise update rule for the membrane potential:
\vspace{-3.5pt}
\begin{equation} \label{neuron-combine}
V[n]=f(V[n-1],I[n]) - V_{th} {s}[n].
\end{equation}
\cref{neuron-combine} is used to define the forward pass of SNNs.

\subsection{Forward Pass}
\label{sec:forward}
\vspace{-3pt}
In this paper, we consider $L$-layer feedforward SNNs with the IF or LIF models. According to \cref{neuron-combine,IF-update}, the spiking dynamics for an SNN with the IF model can be described as:
\vspace{-2pt}
\begin{equation} \label{IF-dynamics}
\mathbf{V}^{i}[n]=\mathbf{V}^{i}[n-1]+V_{t h}^{i-1} \mathbf{W}^{i} \mathbf{s}^{i-1}[n]-V_{t h}^{i} \mathbf{s}^{i}[n],
\end{equation}
where $i=1,2,\cdots,L$ is the layer index, $\mathbf{s}^0$ are the input data to the network, $\mathbf{s}^i$ are the output spike trains of the $i^{\text{th}}$ layer for $i=1,2,\cdots,L$, and $\mathbf{W}^i$ are trainable synaptic weights from the $(i-1)^{\text{th}}$ layer to the $i^{\text{th}}$ layer. Spikes are generated 
according to Eq.~\eqref{discreteb}, and $V_{t h}^{i-1} \mathbf{W}^{i} \mathbf{s}^{i-1}[n]$ are treated as input currents to the $i^{\text{th}}$ layer. Furthermore, the spike thresholds are the same for all IF neurons in one particular layer.
Similarly, according to \cref{neuron-combine,LIF-update}, the spiking dynamics for an SNN with the LIF model can be shown as:
\vspace{-2pt}
\begin{equation} \label{LIF-dynamics} \small
\begin{aligned}
\mathbf{V}^{i}&[n]=\exp({-\frac{\Delta t}{\tau^i}}) \mathbf{V}^{i}[n-1] \\
&+ (1-\exp({-\frac{\Delta t}{\tau^i}})) \frac{V_{t h}^{i-1}}{\Delta t}   \mathbf{W}^{i} \mathbf{s}^{i-1}[n]-V_{t h}^{i} \mathbf{s}^{i}[n],
\end{aligned}
\end{equation}
where $\Delta t$ is set to be a positive number much less than $\tau_i$, and it appears to simplify the analysis on spike representation schemes in \cref{sec:representation}.
In \cref{IF-dynamics,LIF-dynamics}, we only consider fully connected layers. However, other neural network components like convolution, skip connection, and average pooling can also be adopted.
\begin{figure}[h]
	\centering
	\includegraphics[width=0.9\columnwidth]{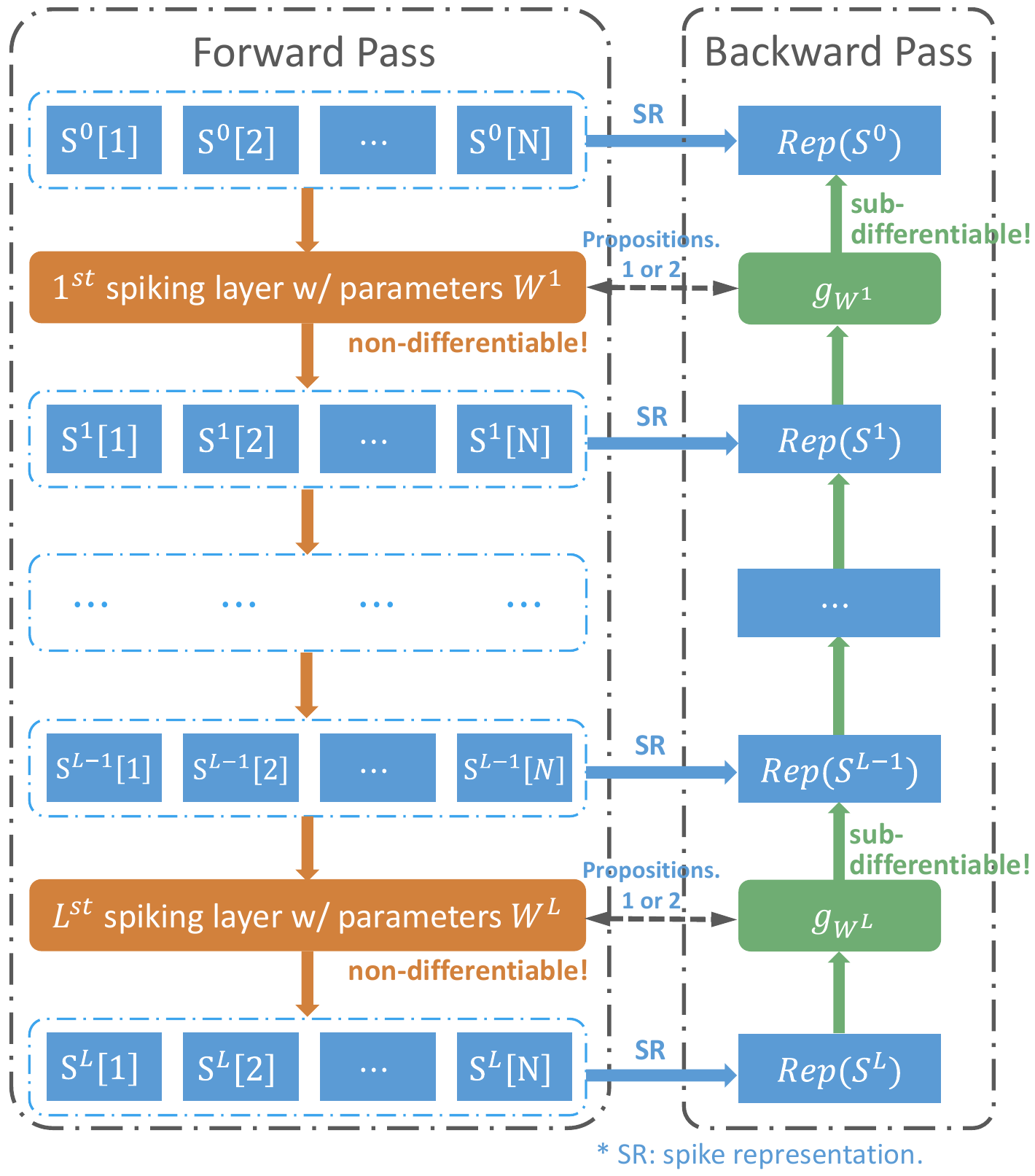} 
	\vspace{-8pt}
	\caption{The pipeline of proposed DSR method. The left part shows the forward computation of an SNN. The right shows the error backpropagation through the sub-differentiable mapping $g_{\mathbf{W}^i}$.}
	\vspace{-14pt}
	\label{fig:pipline}	
\end{figure}

The input $\mathbf{s}^0$ to the SNN can be both neuromorphic data or static data (\eg, images). While neuromorphic data are naturally adapted to SNNs, for static data, we repeatedly apply them to the first layer at each time step \cite{xiao2021ide,zheng2020going,rueckauer2017conversion}. With this method, the first layer can be treated as the spike-train data generator. We use the spike trains $\mathbf{s}^L$ as the output data of the SNN, whose setting is more biologically plausible than prohibiting firing for the last layer and using the membrane potentials as the network output \cite{wu2021training,lee2020enabling}.

\subsection{Spike Representation}
\label{sec:representation}
In this subsection, we show that the forward computation for each layer of an SNN with the IF or LIF neurons can be represented as a sub-differentiable mapping using spike representation as the information carrier. And the spike representation is obtained by (weighted) firing rate coding. Specifically, denoting by $\mathbf{s}^i$ the output spike trains of the $i^{\text{th}}$ layer, the relationship between the SNN and the mapping can be expressed as
\vspace{-2pt}
\begin{equation}{} \label{representation}
	\operatorname{Rep}(\mathbf{s}^i) \approx g_{\mathbf{W}^{i}}(\operatorname{Rep}(\mathbf{s}^{i-1})), \ i=1,2,\cdots,L,
\end{equation}
where $\operatorname{Rep}(\mathbf{s}^i)$ is the spike representation of $\mathbf{s}^i$, $\mathbf{W}^{i}$ are the SNN parameters for the $i^{\text{th}}$ layer, and $g_{\mathbf{W}^{i}}$ is the sub-differentiable mapping also parameterized by $\mathbf{W}^{i}$. Then the SNN parameters $\mathbf{W}^{i}$ can be learned through gradients of $g_{\mathbf{W}^{i}}$. The illustration of the SNN-to-mapping representation is shown in \cref{fig:pipline}.

We first use weighted firing rate coding to derive the formulae of spike representation $\operatorname{Rep}(\mathbf{s}^i)$ and the sub-differentiable mapping $g_{\mathbf{W}^{i}}$ for the LIF model. Then we briefly introduce the formulae for the IF model, which are simple  extensions of those for the LIF model. Training SNNs based on the spike representation schemes is described in \cref{sec:train}.

\vspace{-8pt}
\subsubsection{Spike Representation for the LIF Model}
\label{sec:SRLIF}
We first consider the LIF model defined by Eqs. \eqref{discretea} to \eqref{discretec} and \eqref{LIF-update}. To simplify the notation,
define $\lambda = \exp(-\frac{\Delta t}{\tau})$. 
We further define $\hat{I}[N]=\frac{\sum_{n=1}^{N} \lambda^{N-n} I[n]}{\sum_{n=1}^{N} \lambda^{N-n}}$ as the weighted average input current until the time step $N$, and define $\hat{a}[N]=\frac{V_{th}\sum_{n=1}^{N} \lambda^{N-n} s[n]}{\sum_{n=1}^{N} \lambda^{N-n}\Delta t}$ as the scaled weighted firing rate. Here we treat  $\hat{a}[N]$ as the spike representation of the spike train $\{s[n]\}_{n=1}^N$ for the LIF model.
The key idea is to directly determine the relationship between $\hat{I}[N]$ and $\hat{a}[N]$ using a (sub-)differentiable mapping.

In detail, combining \cref{neuron-combine,LIF-update}, and multiplying the combined equation by $\lambda^{N-n}$, we have
\begin{equation} \small \label{LIF-combine}
\begin{aligned}
\lambda^{N-n} V[n] &= \lambda^{N-n+1} V[n-1] \\
&+ (1-\lambda)\lambda^{N-n}I[n] -\lambda^{N-n}V_{th}s[n].
\end{aligned}
\end{equation}
Summing \cref{LIF-combine} over $n=1$ to $N$, we can get
\begin{equation} \small \label{LIF-sumsd}
\begin{aligned}
V[N] = (1-\lambda)\sum_{n=1}^{N}\lambda^{N-n} I[n] - \sum_{n=1}^{N}\lambda^{N-n} V_{th}s[n].
\end{aligned}
\end{equation}
Dividing \cref{LIF-sumsd} by $\Delta t\sum_{n=1}^{N} \lambda^{N-n}$ and then rearrange the terms, we have
\begin{equation} \label{LIF-approximasdte}
\hat{a}[N]  = \frac{(1-\lambda)\hat{I}[N]}{\Delta t}  - \frac{V[N]}{\Delta t\sum_{n=1}^{N} \lambda^{N-n}}.
\end{equation}
Note that we can further approximate $\frac{1-\lambda}{\Delta t}$ in \cref{LIF-approximasdte} by $\frac{1}{\tau}$, since $\lim\limits_{\Delta t\rightarrow0}\frac{1-\lambda}{\Delta t} = \frac{1}{\tau}$ and we set $\Delta t \ll \tau$. Then we have
\begin{equation} \label{LIF-approximasdte2}
\hat{a}[N]\approx \frac{\hat{I}[N]}{\tau}  - \frac{V[N]}{\Delta t\sum_{n=1}^{N} \lambda^{N-n}}.
\end{equation}

\cref{LIF-approximasdte2} is the basic formula to determine the mapping from $\hat{I}[N]$ to $\hat{a}[N]$. 
Note that in \cref{LIF-approximasdte2}, the term $\frac{V[N]}{\Delta t\sum_{n=1}^{N} \lambda^{N-n}}$ cannot be directly determined only given $\hat{I}[N]$.
However, taking $\hat{a}[N] \in [0,\frac{V_{th}}{\Delta t}]$ into consideration and assuming $V_{th}$ is small, we can ignore the term $\frac{V[N]}{\Delta t\sum_{n=1}^{N} \lambda^{N-n}}$ in \cref{LIF-approximasdte2}, and further approximate $\hat{a}[N]$ as
\vspace{-2pt}
\begin{equation}{} \label{LIF-approximasdte3}
\begin{aligned}
\lim\limits_{N\rightarrow \infty}\hat{a}[N] \approx \operatorname{clamp}\left(\lim\limits_{N\rightarrow \infty}\frac{\hat{I}[N]}{\tau}, 0, \frac{V_{t h}}{\Delta t}\right),
\end{aligned}
\end{equation}
where $\operatorname{clamp}(x,a,b)\triangleq\max (a, \min (x, b))$.
Detailed derivation and mild assumptions for \cref{LIF-approximasdte3} are shown in the Supplementary Materials. Applying \cref{LIF-approximasdte3} to feedforward SNNs with multiple LIF neurons, we have \cref{coro:lif-weight}, which is used to train SNNs. 
\begin{proposition}\label{coro:lif-weight}
	Consider an SNN with LIF neurons defined by \cref{LIF-dynamics}.
	Define $\hat{\mathbf{a}}^0[N]=\frac{\sum_{n=1}^{N} \lambda_i^{N-n} \mathbf{s}^0[n]}{\sum_{n=1}^{N} \lambda_i^{N-n}\Delta t}$ 
	and $\hat{\mathbf{a}}^i[N]=\frac{V_{th}^i\sum_{n=1}^{N} \lambda_i^{N-n} \mathbf{s}^i[n]}{\sum_{n=1}^{N} \lambda_i^{N-n}\Delta t}, \forall i=1,2,\cdots,L$, where $\lambda_i = \exp(-\frac{\Delta t}{\tau^i})$.  Further define sub-differentiable mappings
	\vspace{-2pt}
	$$
	\mathbf{z}^{i}=\operatorname{clamp}\left(\frac{1}{\tau^i}\mathbf{W}^{i} \mathbf{z}^{i-1}, 0, \frac{V_{t h}^{i}}{\Delta t}\right), i=1,2,\cdots,L. 
	$$
	If $\lim\limits_{N\rightarrow \infty}\hat{\mathbf{a}}^i[N] = \mathbf{z}^i$ for $i=0,1,\cdots,L-1$, then $\hat{\mathbf{a}}^{i+1}[N]$ approximates $\mathbf{z}^{i+1}$ when $N \rightarrow\infty$.
\end{proposition}

\vspace{-15pt}
\subsubsection{Spike Representation for the IF Model}
We then consider the IF model defined by Eqs. \eqref{discretea} to \eqref{discretec} and \eqref{IF-update}.
Define $\bar{I}[N]=\frac{1}{N} \sum_{n=1}^{N} I[n]$ as the average input current until the time step $N$, and define $a[N]=\frac{1}{N} \sum_{n=1}^{N} V_{t h} s[n]$ as the scaled firing rate. We treat $a[N]$ as the spike representation of the spike train $\{s[n]\}_{n=1}^N$ for the IF model.
We can use similar arguments shown in \cref{sec:SRLIF} to determine 
the relationship between $\hat{I}[N]$ and $\hat{a}[N]$ as
\vspace{-2pt}
\begin{equation}{} \label{IF-approximate3}
\begin{aligned}
\lim\limits_{N\rightarrow \infty}a[N] = \operatorname{clamp}\left(\lim\limits_{N\rightarrow \infty}\bar{I}[N], 0, V_{t h}\right).
\end{aligned}
\end{equation}
Detailed assumptions and derivation for \cref{IF-approximate3} are shown in the Supplementary Materials.
With \cref{IF-approximate3}, we can have \cref{coro:if} to train feedforward SNNs with the IF model.

\begin{proposition}\label{coro:if}
	Consider an SNN with IF neurons defined by \cref{IF-dynamics}. Define $\mathbf{a}^0[N]=\frac{1}{N} \sum_{n=1}^{N} \mathbf{s}^{0}[n]$ and $\mathbf{a}^i[N]=\frac{1}{N} \sum_{n=1}^{N} V_{t h}^{L} \mathbf{s}^{i}[n], \forall i=1,2,\cdots,L$. Further define sub-differentiable mappings:
	$$
	\mathbf{z}^{i}=\operatorname{clamp}\left(\mathbf{W}^{i} \mathbf{z}^{i-1}, 0, V_{t h}^{i}\right), i=1,2,\cdots,L. 
	$$
	If $\lim\limits_{N\rightarrow \infty}\mathbf{a}^0[N] = \mathbf{z}^0$, then $\lim\limits_{N\rightarrow \infty}\mathbf{a}^i[N]=\mathbf{z}^{i}$.
\end{proposition}

\subsection{Differentiation on Spike Representation}
\label{sec:train}
In this subsection, we use the spike representation for the IF and the LIF models to drive the backpropagation training algorithm for SNNs, based on \cref{coro:if,coro:lif-weight}. And the illustration can be found in \cref{fig:pipline}.

Define the spike representation operator $\operatorname{r}(\cdot)$ with spike train $s=(s[1],\cdots,s[N])$ as input, such that $\operatorname{r}(s)=\frac{1}{N}\sum_{n=1}^{N} V_{t h} s[n]$ for the IF model, and $\operatorname{r}(s)=\frac{V_{th}\sum_{n=1}^{N} \lambda^{N-n} s[n]}{\sum_{n=1}^{N} \lambda^{N-n}\Delta t}$ for the LIF model, where $\lambda = \exp(-\frac{\Delta t}{\tau})$. With the spike representation, we define the final output of the SNN as $\mathbf{o}^L=\operatorname{r}(\mathbf{s}^L)$, where $\mathbf{s}^L$ is the output spike trains from the last layer and $\operatorname{r}(\cdot)$ is defined element-wise. We use cross-entropy as the loss function $\ell$.

The proposed DSR method backpropagates the gradient of error signals based on the representation of spike trains in each layer, $\mathbf{o}^i=\operatorname{r}(\mathbf{s}^i)$, where $i=1,2,\cdots,L$ is the layer index. By applying chain rule, the required gradient $\frac{\partial \ell}{\partial \mathbf{W}^i}$ can be computed as
\begin{equation}
\frac{\partial \ell}{\partial \mathbf{W}^i} = \frac{\partial \ell}{\partial \mathbf{o}^i}  \frac{\partial \mathbf{o}^i}{\partial \mathbf{W}^i} , \quad
\frac{\partial \ell}{\partial \mathbf{o}^i} = \frac{\partial \ell}{\partial \mathbf{o}^{i+1}}  \frac{\partial \mathbf{o}^{i+1}}{\partial \mathbf{o}^i} ,
\end{equation}
where $\frac{\partial\mathbf{o}^{i+1}}{\partial \mathbf{o}^i}$ and $\frac{\partial \mathbf{o}^i}{\partial \mathbf{W}^i}$ can be computed with \cref{coro:if,coro:lif-weight}. Specifically, from \cref{sec:representation}, we have 
\begin{equation} \small \label{representation-all}
\mathbf{o}^i = \operatorname{r}(\mathbf{s}^i) \approx \operatorname{clamp}(\mathbf{W}^i\operatorname{r}(\mathbf{s}^{i-1}),0,b_i), \ i=1,2,\cdots,L,
\end{equation}
where $b_i=V_{th}^i$ for the IF model, and $b_i=\frac{V_{th}^i}{\Delta t}$ for the LIF model. Therefore, we can calculate $\frac{\partial\mathbf{o}^{i+1}}{\partial \mathbf{o}^i}$ and $\frac{\partial \mathbf{o}^i}{\partial \mathbf{W}^i}$ based on \cref{representation-all}. The pseudocode of the proposed DSR method can be found in the Supplementary Materials.

With the proposed DSR method, we avoid two common problems in SNN training. First, this method does not require backpropagation through the temporal domain, improving the training efficiency when compared with the BPTT type methods, especially when the number of time steps is not ultra-small. Second, this method does not need to handle the non-differentiability of spike functions, since the signals are backpropagated through sub-differentiable mapping. Although there exists representation error due to finite time steps, we can reduce it, as described in \cref{sec:reduce-error}.


\vspace{-2pt}
\section{Reducing Representation Error}
\label{sec:reduce-error}
\vspace{-2pt}
\cref{coro:if,coro:lif-weight} show that the (weighted) firing rate can gradually estimate or converge to the output of a sub-differentiable mapping. And \cref{sec:train} shows that we can train SNNs by backpropagation using spike representation. However, in practice we want to simulate SNNs with only a small number of time steps, for the sake of low energy consumption. The low latency will further introduce representation error that hinders effective training. In this subsection, we study the representation error and propose to train the spike threshold and introduce a new hyperparameter for the neural models to reduce the error.

The representation error $e_r$ can be decomposed as $e_r=e_q+e_d$, where $e_q$ is the ``quantization error'' and $e_d$ is the ``deviation error''. The quantization error $e_q$ exists due to the imperfect precision of the firing rate, when assuming the same input currents at all time steps. For example, it can only take value in the form $\frac{n}{N}$ for the IF neuron, where $n\in\mathbb{N}$ and $N$ is the number of time steps. And the deviation error $e_q$ exists due to the inconsistency of input currents at different time steps. For example, when the average input current is $0$, the output firing rate is supposed to be 0; however, it can be significantly larger than $0$ if the input currents are positive during the first few time steps. 

\begin{figure}[h]
	\centering
	\includegraphics[width=0.8\columnwidth]{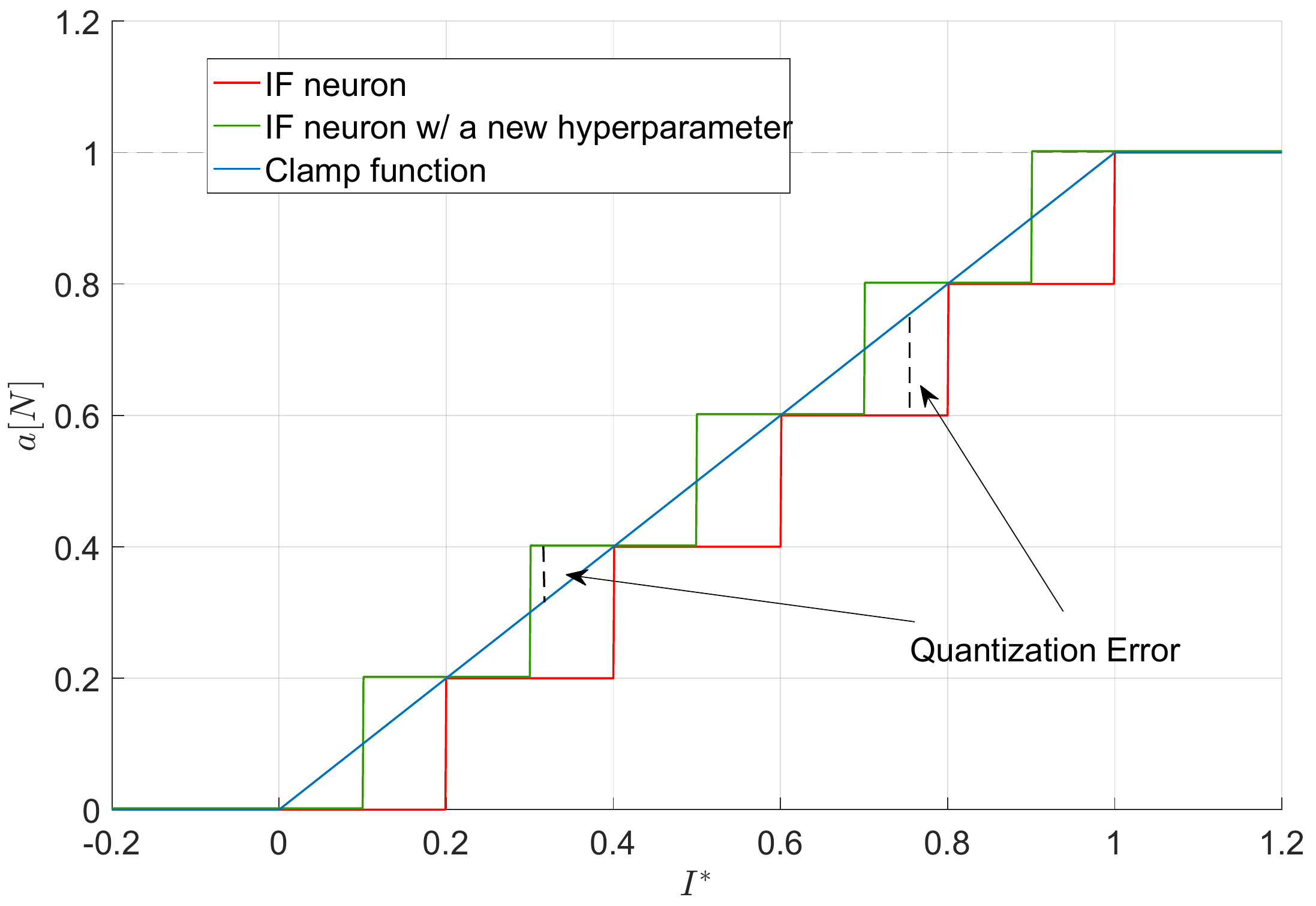} 
	\vspace{-4pt}
	\caption{ The firing rate of the IF neuron approximates the clamp function given unchanged input currents. Spike generating of IF neurons w/ or w/o introducing a new hyperparameter for spiking neurons is controlled by Eqs.~\eqref{discreteb} and \eqref{modify-reset}, respectively. Here we set the threshold $\theta=1$ and the latency $N=5$.}
	\vspace{-12pt}
	\label{fig:IF-ReLU}	
\end{figure}

From the statistical perspective, the expectation for $e_d$ is 0, assuming \text{i.i.d.} input currents at different time steps. Therefore, the ``deviation error'' $e_d$ will not affect training too much when using stochastic optimization algorithms. Next, we dig into $e_q$ with the IF model and then propose methods to reduce it. Similar arguments can be derived for the LIF model. When given unchanged input currents $I^*$ at all time steps to the IF neuron, the scaled average firing rate $a[N]=\frac{1}{N} \sum_{n=1}^{N} V_{t h} s[n]$ can be determined as 
\begin{equation} \label{constant-represent}
a[N]=\frac{V_{th}}{N} \cdot \operatorname{clamp}\left(\left\lfloor\frac{N I^*}{\theta}\right\rfloor, 0, N\right),
\end{equation}
shown as the red curve in Fig.~\ref{fig:IF-ReLU}, where $\lfloor\cdot\rfloor$ is the floor rounding operator. Inspired by \cref{constant-represent}, we propose two methods to reduce the quantization error.

\vspace{-7pt}
\paragraph{Training the Spike Threshold}

From \cref{fig:IF-ReLU} and \cref{constant-represent}, we observe that using small spike thresholds can reduce the quantization error. However, it also weakens the approximation capacity of the SNNs, since the scaled (weighted) firing rate will be in a small range then. Inspired by activation clipping methods for training quantized neural networks \cite{choi2018pact}, in this paper, we treat the spike threshold of each layer as parameters to be trained, and include an L2-regularizer for the thresholds in the loss function to balance the tradeoff between quantization error and approximation capacity.
To train the spike thresholds using backpropagation, we calculate the gradients with respect to them based on the spike representation introduced in \cref{sec:representation}. For example, using \cref{IF-approximate3}, for one IF neuron with average input current $I^*$ and steady scaled firing rate $a^*$, we have
\begin{equation}
\frac{\partial a^*}{\partial V_{th}}=\left\{
\begin{array}{l}
1, \quad \text{if} \  I^*>V_{th},  \\
0, \quad \text{otherwise}.
\end{array}\right.
\end{equation}
Then we can calculate the gradient of the loss function with respect to the threshold by the chain rule. A similar calculation applies to LIF neurons. In practice, since we use mini-batch optimization methods to train SNNs, the gradient for each threshold is proportional to the batch size by the chain rule. Thus, we scale the gradient regarding different batch sizes and spiking neural models.

\vspace{-10pt}
\paragraph{Introducing a new hyperparameter for the neural models} 
We can introduce a new hyperparameter for spiking neurons to control the neuron firing to reduce the quantization error. Formally, we change Eq. \eqref{discreteb} to
\begin{equation}
s[n]=H(U[n]-\alpha V_{th}) \label{modify-reset}
\end{equation}
to get a new firing mechanism, where $\alpha\in[0,1]$ is a hyperparameter. For the IF model with the new firing mechanism and $\alpha=0.5$, using the same notation as in \cref{constant-represent}, the scaled firing rate becomes
\begin{equation} \label{constant-represent-modified}
a[N]=\frac{V_{th}}{N} \cdot \operatorname{clamp}\left(\left[\frac{N I^*}{\theta}\right], 0, N\right),
\end{equation}
shown as the green curve in Fig.~\ref{fig:IF-ReLU}, where $[\cdot]$ is the rounding operator. From \cref{fig:IF-ReLU}, we can see that the maximum absolute quantization error is halved when using this mechanism. Furthermore, since $\alpha=0.5$ makes the average absolute quantization error minimized, $\alpha=0.5$ is the best choice for the IF model. On the other hand, for the LIF model, the best choice for $\alpha$ changes when setting different latency $N$, so we choose different $\alpha$ in our experiments to minimize the average absolute quantization error.

\begin{table*}[h]
	\caption{Performance on CIFAR-10, CIFAR-100, ImageNet, and DVS-CIFAR10. For the first three datasets, we categorize the methods into 4 classes: ANN, the ANN-to-SNN method, the direct training method, and our proposed method. Different types of methods are separated by horizontal lines. We bold the best result for the LIF model, and underline the best result for the IF model.}
	\label{table:compare}
	\centering
	\vspace{-6pt}
	\begin{threeparttable}
		\begin{tabular}{c|lcccc}
			\toprule[1.08pt] & Method & Network & Neural Model & Time Steps  & Accuracy \\
			\midrule[1.08pt]
			
			\multirow{12}*{\rotatebox{90}{CIFAR-10}} 
			& ANN \tnote{1} & PreAct-ResNet-18 & / & /  & $95.41\%$  \\
			\cline{2-6}
			& ANN-to-SNN\cite{deng2021optimal} & ResNet-20 & IF & 128  & $93.56\%$  \\
			& ANN-to-SNN\cite{han2020rmp} & VGG-16 & IF & 2048  & $93.63\%$  \\
			&ANN-to-SNN\cite{yan2021near} & VGG-like & IF & 600 & $94.20\%$ \\
			\cline{2-6}
			&Tandem Learning\cite{wu2021tandem} & CIFARNet & IF & 8 & $90.98\%$  \\
			& ASF-BP\cite{wu2021training} & VGG-7 & IF & 400  & $91.35\%$  \\
			&STBP\cite{wu2018spatio} & CIFARNet & LIF & 12  & $90.53\%$  \\
			&IDE\cite{xiao2021ide} & CIFARNet-F & LIF & 100 &  $92.52\%\pm0.17\%$ \\
			& STBP-tdBN\cite{zheng2020going} & ResNet-19 & LIF & 6  & $93.16\%$  \\
			&TSSL-BP\cite{zhang2020temporal} & CIFARNet & LIF w/ synaptic model & 5 & $91.41\%$  \\
			\cline{2-6}
			&\multirow{2}{*}{DSR (ours)} & PreAct-ResNet-18 & IF & 20 & $\underline{95.24\%\pm0.17\%}$ \\
			&& PreAct-ResNet-18 & LIF & 20  & $\mathbf{95.40\%\pm0.15\%}$  \\
			\midrule[1.08pt]
			
			\multirow{9}*{\rotatebox{90}{CIFAR-100}} 
			& ANN \tnote{1} & PreAct-ResNet-18 & / & /  & $78.12\%$  \\ 
			\cline{2-6}
			&ANN-to-SNN\cite{deng2021optimal} & ResNet-20 & IF & 400-600 & $69.82\%$ \\
			&ANN-to-SNN\cite{han2020rmp} & VGG-16 & IF & 768 & $70.09\%$ \\
			&ANN-to-SNN\cite{yan2021near} & VGG-like & IF & 300 & $71.84\%$ \\
			\cline{2-6}
			&Hybrid Training\cite{rathi2019enabling} & VGG-11 & LIF & 125  & $67.84\%$  \\
			&DIET-SNN\cite{rathi2020diet} & VGG-16 & LIF & 5 & $69.67\%$  \\
			&IDE\cite{xiao2021ide} & CIFARNet-F & LIF & 100 & $73.07\%\pm0.21 \%$ \\
			\cline{2-6}
			&\multirow{2}{*}{DSR (ours)} & PreAct-ResNet-18 & IF & 20 & $\underline{78.20\%\pm0.13\%}$  \\
			&& PreAct-ResNet-18 & LIF & 20  & $\mathbf{78.50\%\pm0.12\%}$ \\
			\midrule[1.08pt]
			
			\multirow{8}*{\rotatebox{90}{ImageNet}} 
			& ANN \tnote{1} & PreAct-ResNet-18 & / & /  & $70.79\%$  \\
			\cline{2-6}
			&ANN-to-SNN\cite{sengupta2019going} & ResNet-34 & IF & 2000 & $65.47\%$ \\
			&ANN-to-SNN\cite{han2020rmp} & ResNet-34 & IF & 4096 & $\underline{69.89\%}$ \\
			\cline{2-6}
			&Hybrid training\cite{rathi2019enabling} & ResNet-34 & LIF & 250 &  $61.48\%$ \\
			&STBP-tdBN\cite{zheng2020going} & ResNet-34 & LIF & 6 & $63.72 \%$  \\
			&SEW ResNet\cite{fang2021sew} & SEW ResNet-34 & LIF & 4  & $67.04\%$ \\
			&SEW ResNet\cite{fang2021sew} & SEW ResNet-18 & LIF & 4  & $63.18\%$ \\
			\cline{2-6}
			&DSR (ours) & PreAct-ResNet-18 & IF & 50  & $\underline{67.74\%}$  \\
			\midrule[1.08pt]
			
			\multirow{7}*{\rotatebox{90}{DVS-CIFAR10}} 
			& ASF-BP\cite{wu2021training} & VGG-7 & IF & 50  & $62.50\%$  \\
			&Tandem Learning\cite{wu2021tandem} & 7-layer CNN & IF & 20 & $65.59\%$  \\
			& STBP\cite{wu2019direct} & 7-layer CNN & LIF & 40  & $60.50\%$  \\
			& STBP-tdBN\cite{zheng2020going} & ResNet-19 & LIF & 10  & $67.80\%$  \\
			&Fang \etal \cite{fang2021incorporating} & 7-layer CNN & LIF & 20 &  $74.80\%$ \\
			\cline{2-6}
			&\multirow{2}{*}{DSR (ours)} & VGG-11 & IF & 20 & $\underline{75.03\%\pm0.39\%}$ \\
			&& VGG-11 & LIF & 20  & $\mathbf{77.27\%\pm0.24\%}$  \\
			\bottomrule[1.08pt]
		
		\end{tabular}
		\begin{tablenotes}
			\footnotesize
			\item[1] Self-implemented results for ANN.
		\end{tablenotes}
	\end{threeparttable}
	\vspace{-12pt}
\end{table*}


\vspace{-3pt}
\section{Experiments}
\label{sec:experiments}
\vspace{-3pt}
We first evaluate the proposed DSR method and compare it with other works on visual object recognition benchmarks, including CIFAR-10, CIFAR-100, ImageNet, and DVS-CIFAR10. We then demonstrate the effectiveness of our method when the number of time steps becomes smaller and smaller, or the network becomes deeper and deeper. We also test the effectiveness of the methods for reducing representation error. Please refer to the Supplementary Materials for experiment details. Our code is available at \url{https://github.com/qymeng94/DSR}.

\subsection{Comparison to the State-of-the-Art}
\vspace{-2pt}
The comparison on CIFAR-10, CIFAR-100, ImageNet, and DVS-CIFAR10 is shown in \cref{table:compare}.

For the CIFAR-10 and the CIFAR-100 datasets, we use pre-activation ResNet-18 \cite{he2016identity} as the network architecture. 
\cref{table:compare} shows that the proposed DSR method outperforms all other methods on CIFAR-10 and the CIFAR-100 with 20 time steps for both the IF and the LIF models, based on 3 runs of experiments. Especially, our method achieves accuracies that are 5\%-10\% higher on CIFAR-100 when compared to others. Furthermore, the obtained SNNs have similar or even better performance compared to ANNs with the same network architectures. Although some direct training methods use smaller time steps than ours, our method can also achieve better performances than others when the number of time steps $N=15,10,$ and $5$, as shown in \cref{fig:timestep}.

For the ImageNet dataset, we also use the pre-activation ResNet-18 network architecture. To accelerate training, we adopt the hybrid training technique \cite{rathi2019enabling,rathi2020diet}. And considering the data complexity and the 1000 classes, we use a moderate number of time steps to achieve satisfactory results. Our proposed method can outperform the direct training methods even if they use larger network architectures. Although some ANN-to-SNN methods have better accuracy, they use much more time steps than ours.

For the neuromorphic DVS-CIFAR10, we adopt the VGG-11 architecture \cite{simonyan2014very} and conduct 3 runs of experiments for each neural model. It can be found in \cref{table:compare} that the proposed method outperforms other SOTA methods with low latency using both the IF and the LIF models.

\vspace{-3pt}
\subsection{Model Validation and Ablation Study}
\begin{figure}[h]
	\centering
		\begin{subfigure}{0.49\linewidth}
			\hspace{-2em}
			\includegraphics[width=1.13\columnwidth]{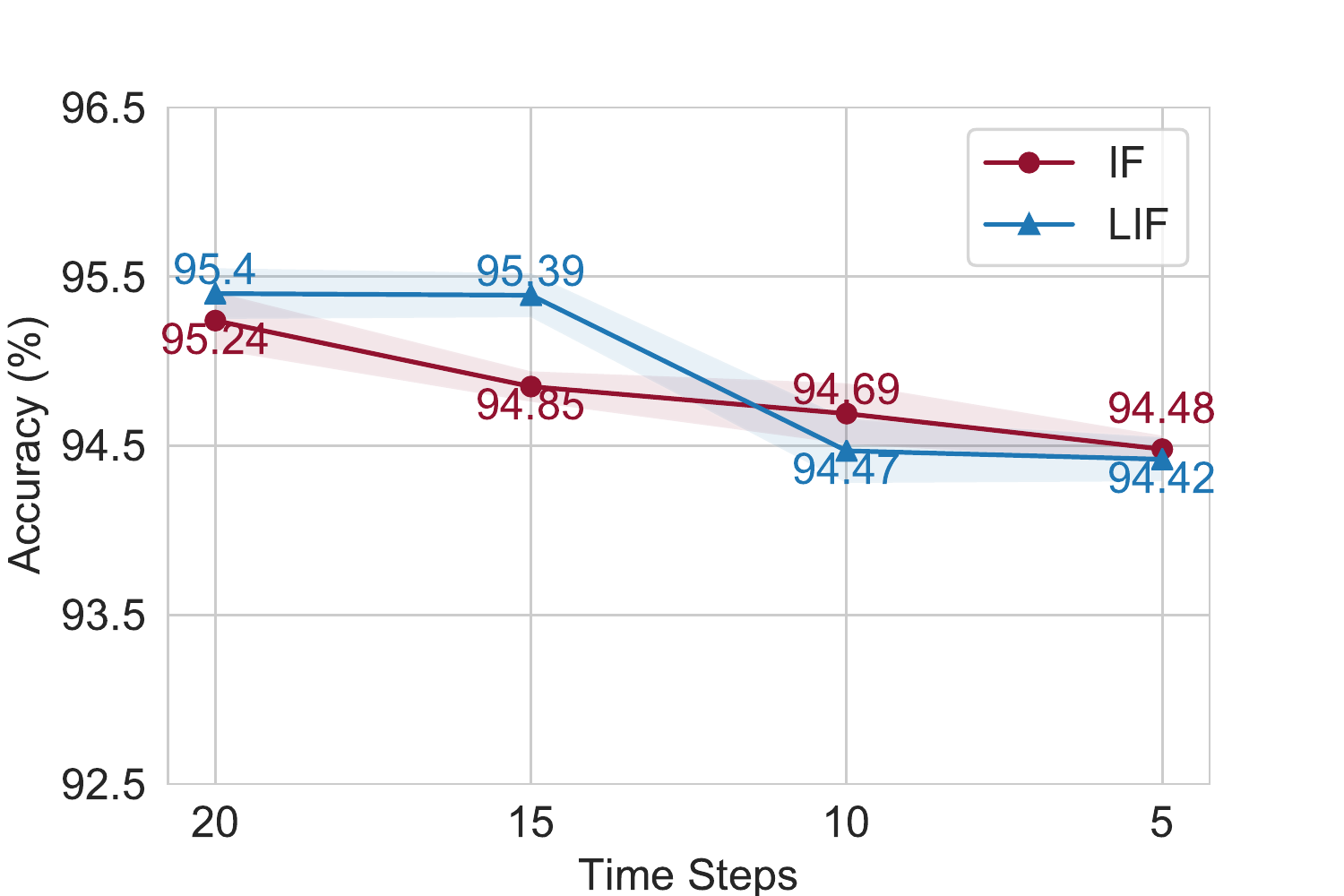} 
			\caption{}
			\label{fig:timestep}
		\end{subfigure}
		\begin{subfigure}{0.49\linewidth}
			\includegraphics[width=1.13\columnwidth]{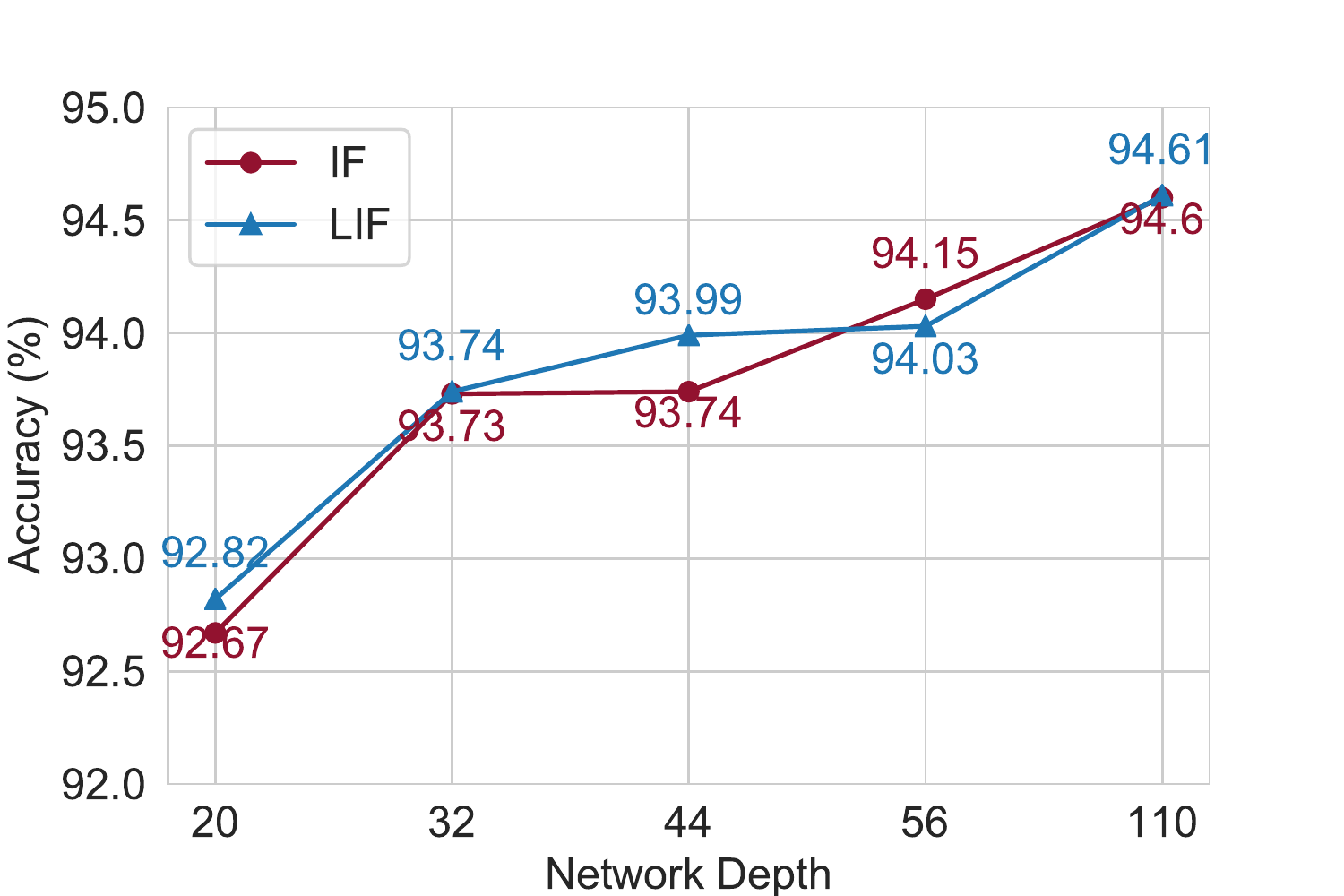} 
			\caption{}
			\label{fig:goingdeeper}
		\end{subfigure}
	\vspace{-10pt}
	\caption{(a) Accuracies achieved by DSR method on CIFAR-10 with low latency. The PreAct-ResNet-18 network architecture is used. Results are based on 3 runs of experiments. (b) Accuracies achieved by DSR method on CIFAR-10 with different network architectures. The number of time steps is 20. Narrow network architectures are used.}
	\label{conversion-exper}
	\vspace{-18pt}
\end{figure}

\vspace{-2pt}
\paragraph{Effectiveness of the Proposed Method with Low Latency} We validate that the proposed method can achieve competitive performances even with ultra-low latency, as shown in \cref{fig:timestep}. Each model is trained from scratch. From 20  to 5 time steps, our models only suffer from less than $1\%$ accuracy drop. The results for 5 time steps also outperform other SOTA shown in \cref{table:compare}. More training details can be found in the Supplementary Materials.

\vspace{-12pt}
\paragraph{Effectiveness of the Proposed Method with Deep Network Structure} Many SNN learning methods cannot adapt to deep network architectures, limiting the potential of SNNs. The reason is that the error for gradient approximation or ANN-to-SNN conversion will accumulate through layers, or the methods are computationally expensive for large-scale network structures. In this part, we test the proposed method on CIFAR-10 using pre-activation ResNet with different depths, namely 20, 32, 44, 56, 110 layers. Note that the channel size is smaller than the PreAct-ResNet-18, since the deep networks with large channel size as in PreAct-ResNet-18 perform not much better and are harder to train even for ANNs. More details about network architectures can be found in the Supplementary Materials. Results are shown in \cref{fig:goingdeeper}. The figure shows that our method is effective on deep networks (\textgreater 100 layers), and performs better with deeper network structures. This indicates the great potential of our method to achieve more advanced performance when using very deep networks.

\vspace{-12pt}
\paragraph{Ablation Study on Methods to Reduce Representation Error} We conduct the ablation study on the representation error reduction methods, namely training the threshold and introducing a new hyperparameter for the neural models.
The models are trained on CIFAR-10 with PreAct-ResNet-18 structure and 20 time steps, and the results are shown in \cref{table:ablation}. The experiments imply that the representation error significantly hinders training and also demonstrate the superiority of the two methods to reduce the representation error. Note that
the threshold training method also helps stabilize training, since the results become unstable for large thresholds without this method (\eg, the standard deviation is $1.84\%$ when $V_{th}=6$). Furthermore, 
the average accuracy of not using both methods is better than the one of only using the firing mechanism modification, maybe due to the instability of the results when $V_{th}$ is large.

\vspace{-6pt}
\begin{table}[h!tp]
	\caption{Ablation study on the representation error reduction methods on CIFAR-10. The PreAct-ResNet-18 architecture with the IF model is used, and results are based on 3 runs of experiments. `F' means firing mechanism modification, and `T' means threshold training.}
	\vspace{-6pt}
	\label{table:ablation}
	\centering
	\begin{threeparttable}
		\begin{tabular}{cc}
			\hline  Setting & Accuracy \\
			\hline 
			DSR, init. $V_{th}=6$  &  $95.24\%\pm0.17\%$ \\
			DSR w/o F, init. $V_{th}=6$ &  $92.88\%\pm0.25\%$  \\
			DSR w/o T, $V_{th}=6$ & $90.45\%\pm1.84\%$\\
			DSR w/o T, $V_{th}=2$ & $90.47\%\pm0.12\%$\\
			DSR w/o F\&T, $V_{th}=6$ & $92.59\%\pm0.81\%$ \\
			\hline
		\end{tabular}
	\end{threeparttable}
	\vspace{-10pt}
\end{table}

\vspace{-5pt}
\section{Conclusion and Discussions}
\label{sec:conclusion}
\vspace{-2pt}
In this work, we show that the forward computation of SNNs can be represented as some sub-differentiable mapping. Based on the SNN-to-mapping representation, we propose the DSR method to train SNNs that avoids the non-differentiability problem in SNN training and does not require backpropagation through the temporal domain. We also analyze the representation error due to the small number of time steps, and propose to train the thresholds and introduce a new hyperparameter for the IF and LIF models to reduce the representation error. With the error reduction methods, we can train SNNs with low latency by the DSR method. Experiments show that the proposed method could achieve SOTA performance on mainstream vision tasks, and show the effectiveness of the method when dealing with ultra-low latency or very deep network structures.

{\bf Societal impact and limitations.} As for societal impact, there is no direct negative societal impact since this work only focuses on training SNNs. In fact, the development of high-performance low-latency SNNs allows SNNs to replace ANNs in some real-world tasks. This replacement will alleviate the huge energy consumption by ANNs and reduce carbon dioxide emissions.
As for limitations, the DSR method may suffer from a certain degree of performance drop when the latency is extremely low (\eg, with only 2 or 3 time steps), since the method requires relatively accurate spike representation to conduct backpropagation.

\vspace{-5pt}
\section*{Acknowledgment}
\vspace{-3pt}
We thank Jiancong Xiao and Zeyu Qin for useful discussion. We thank Jin Wang for pointing out typos. The work of Z.-Q. Luo was supported by the National Natural Science Foundation of China under Grant 61731018, and the Guangdong Provincial Key Laboratory of Big Data Computation Theories and Methods. Z. Lin was supported by the NSF China (No. 61731018), NSFC Tianyuan Fund for Mathematics (No. 12026606), Project 2020BD006 supported by PKU-Baidu Fund, and Qualcomm. The work of Yisen Wang was partially supported by the National Natural Science Foundation of China under Grant 62006153, and Project 2020BD006 supported by PKU-Baidu Fund.

{\small
\bibliographystyle{ieee_fullname}
\bibliography{arxiv}
}

\appendix
\newpage

\section{Details about Spike Representation}
\subsection{Derivation for \cref{LIF-approximasdte3}}
\label{sec:lif-derive}
In this subsection, we consider the LIF model defined by Eqs. \eqref{discretea} to \eqref{discretec} and \eqref{LIF-update}, and derive \cref{LIF-approximasdte3} from \cref{LIF-approximasdte2} in the main content under mild assumptions.

From the main content, we have derived that 
\begin{equation} \label{LIF-1}
\hat{a}[N]\approx \frac{\hat{I}[N]}{\tau}  - \frac{V[N]}{\Delta t\sum_{n=1}^{N} \lambda^{N-n}},
\end{equation}
as shown in \cref{LIF-approximasdte2}. Since the LIF neuron is supposed to fire no or very few spikes when $\frac{\hat{I}[N]}{\tau}<0$ and fire almost always when $\frac{\hat{I}[N]}{\tau}>\frac{V_{t h}}{\Delta t}$, we can separate the accumulated membrane potential $V^[N]$ into two parts: one part $V^-[N]$ represents the ``exceeded'' membrane potential that does not contribute to spike firing, and the other part $V^+[N]$ represents the ``remaining'' membrane potential. In detail, the ``exceeded'' membrane potential $V^-[N]$ can be calculated as
\begin{equation} \small
V^-[N]=\left\{
\begin{array}{l}
(\Delta t\sum_{n=1}^{N} \lambda^{N-n}) \frac{\hat{I}[N]}{\tau}, \quad  \quad \quad \quad \quad \ \ \  \frac{\hat{I}[N]}{\tau}<0,  \\
(\Delta t\sum_{n=1}^{N} \lambda^{N-n})(\frac{\hat{I}[N]}{\tau} - \frac{V_{t h}}{\Delta t}),    \quad  \frac{\hat{I}[N]}{\tau}>\frac{V_{t h}}{\Delta t},\\
0, \quad \quad \quad \quad \quad \quad \quad \quad \quad \quad \quad \quad \quad \quad \ \ \text{otherwise}.
\end{array}\right.
\end{equation}
And the ``remaining'' membrane potential can be calculated as $V^+[N]=V[N]-V^-[N]$. With the decomposition of membrane potential $V[N]$ and the fact that $\Delta t\sum_{n=1}^{N} \lambda^{N-n} = \tau$ when $N\rightarrow\infty$ and $\Delta t \rightarrow 0$,
we can further approximate $\hat{a}[N]$ from \cref{LIF-1} as 
\begin{equation} \small \label{LIF-3}
\begin{aligned}
\lim\limits_{\substack{N\rightarrow \infty }}\hat{a}[N] \approx\lim\limits_{\substack{N\rightarrow \infty}}\operatorname{clamp}\left(\frac{\hat{I}[N]}{\tau} - \frac{V^+[N]}{\tau}, 0, \frac{V_{t h}}{\Delta t}\right),
\end{aligned}
\end{equation}
if the limit of right hand side exists.

Now we want to find the condition to ignore the term $\frac{V^+[N]}{\tau}$ in \cref{LIF-3}. In the case $V^-[N]\ne 0$, the magnitude of membrane potential $|V(n)|$ would gradually increase with time. After introducing $V^-$, the ``remaining'' membrane potential $V^+[N]$ typically does not diverge over time. In fact, $V^+[N]$ is typically bounded in $[0,V_{th}]$ when $N\rightarrow\infty$, except in the extreme case when the input current at different time steps distributes extremely unevenly. So we can just assume that $V^+[N]\in[0,V_{th}]$. Furthermore, if we set a significantly smaller threshold $V_{th}$ compared to the magnitude of $\hat{I}[N]$, the term $\frac{V^+[N]}{\tau}$ can be ignored. Then from \cref{LIF-3}, we have
\begin{equation}{} \label{LIF-approximate3}
\begin{aligned}
\lim\limits_{N\rightarrow \infty}\hat{a}[N] \approx \operatorname{clamp}\left(\lim\limits_{N\rightarrow \infty}\frac{\hat{I}[N]}{\tau}, 0, \frac{V_{t h}}{\Delta t}\right).
\end{aligned}
\end{equation}
That is ,we can approximate $\hat{a}[N]$ by $\operatorname{clamp}\left(\frac{\hat{I}[N]}{\tau}, 0, \frac{V_{t h}}{\Delta t}\right)$, with an approximation error bounded by $\frac{V_{th}}{\tau}$ when $N\rightarrow\infty$.

In summary, we derive \cref{LIF-approximasdte3} from \cref{LIF-approximasdte2} in the main content under following mild conditions:
\begin{itemize}
	\item[1.] The LIF neuron fires no or finite spikes as $N\rightarrow\infty$ when $\frac{\hat{I}[N]}{\tau}<0$. And the LIF neuron does not fire only at a finite number of time steps as $N\rightarrow\infty$ when $\frac{\hat{I}[N]}{\tau}>\frac{V_{t h}}{\Delta t}$.
	\item[2.] $V^+[N]\in[0,V_{th}]$.
\end{itemize}

\subsection{Derivation for \cref{IF-approximate3}}
In this subsection, we consider the IF model defined by Eqs. \eqref{discretea} to \eqref{discretec} and \eqref{IF-update} and derive \cref{IF-approximate3} in the main content under mild assumptions.

Combining Eqs. \eqref{IF-update} and \eqref{discretec}, and taking the summation over $n=1$ to $N$, we can get
\begin{equation}{} \label{IF-sum}
V[N] - V[0] = \sum_{n=1}^{N} I[n] - V_{th}\sum_{n=1}^{N} s[n].
\end{equation}
Define the scaled firing rate until the time step $N$ as $a[N]=\frac{1}{N} \sum_{n=1}^{N} V_{t h} s[n]$, and the average input current as $\bar{I}[N]=\frac{1}{N} \sum_{n=1}^{N} I[n]$. Dividing \cref{IF-sum} by $N$, we have
\begin{equation}{} \label{IF-approximate}
a[N] = \bar{I}[N] - \frac{V[N]}{N}.
\end{equation}
Using similar arguments appeared in Section \ref{sec:lif-derive}, we can get 
\begin{equation} \small \label{IF-approximate2}
\begin{aligned}
\lim\limits_{N\rightarrow \infty}a[N] = \lim\limits_{N\rightarrow \infty} \operatorname{clamp}\left(\bar{I}[N]- \frac{V^+[N]}{N}, 0, V_{t h}\right),
\end{aligned}
\end{equation}
if the limit of right hand side exists. Here $V^+[N]=V[N]-V^-[N]$ and 
\begin{equation}{} \small
V^-[N]=\min\left(\max\left(\sum_{n=1}^{N} I[n]-NV_{th},0\right),\sum_{n=1}^{N} I[n]\right).
\end{equation}
Similar to the LIF model, the ``remaining'' membrane potential $V^+[N]$ for the IF model is typically bounded in $[0,V_{th}]$ when $N\rightarrow\infty$, except in the extreme case. For example, consider $\bar{I}[N]>0$ and  the input current is non-zero only at the last $\frac{N}{2}$ time steps, then $V^+[N]$ will be inconsistently large and will be unbounded when $N\rightarrow\infty$. However, this extreme case will not happen in SNN computation for normal input data.
Therefore, we assume $V^+[N]\in[0,V_{th}]$, and can get
\begin{equation}{} \label{IF-3}
\begin{aligned}
\lim\limits_{N\rightarrow \infty}a[N] = \operatorname{clamp}\left(\lim\limits_{N\rightarrow \infty}\bar{I}[N], 0, V_{t h}\right).
\end{aligned}
\end{equation}
if the limit of  $\bar{I}[N]$ exists.

In summary, we derive \cref{IF-approximate3} in the main content under following mild conditions:
\begin{itemize}
	\item[1.] The IF neuron fires no or finite spikes as $N\rightarrow\infty$ when ${\bar{I}[N]}<0$. And the IF neuron does not fire only at a finite number of time steps as $N\rightarrow\infty$ when ${\bar{I}[N]}>{V_{t h}}$.
	\item[2.] $V^+[N]\in[0,V_{th}]$.
\end{itemize}


\section{Pseudocode of the Proposed DSR Method}
We present the pseudocode of one iteration of SNN training with the DSR method in \cref{algo} for better illustration.

\begin{algorithm}[h] 
	\caption{One iteration of SNN training with the proposed DSR method.}
	\label{algo}
	\begin{algorithmic}[1] 
		\Require Time steps $N$; Network depth $L$; Network parameters $\mathbf{W}^L, \cdots, \mathbf{W}^L, V_{th}^1,\cdots,V_{th}^L$; Input data $\mathbf{x}$; Label $\mathbf{y}$; Other hyperparameters.
		\Ensure Trained network parameters $\mathbf{W}^L, \cdots, \mathbf{W}^L$, $V_{th}^1,\cdots,V_{th}^L$.
		\item[\textbf{Forward:}]
		\For {$n=1,2,\cdots,N$}
		\For {$i=1,2,\cdots,L$}
		\If {the IF model is used}
		\State Calculate $\mathbf{s}^i[n]$ by \cref{IF-dynamics};
		\ElsIf {the LIF model is used} 
		\State Calculate $\mathbf{s}^i[n]$ by \cref{LIF-dynamics};
		\EndIf
		
		\If {$n=N$}
		
		\If {the IF model is used}
		\State $\mathbf{o}^i=\frac{1}{N}\sum_{n=1}^{N} V_{t h} \mathbf{s}[n]$;
		\ElsIf {the LIF model is used} 
		\State $\mathbf{o}^i=\frac{V_{th}\sum_{n=1}^{N} \lambda^{N-n} s[n]}{\sum_{n=1}^{N} \lambda^{N-n}\Delta t}$;
		\EndIf
		
		\EndIf
		\EndFor
		\EndFor
		\State Calculate the loss $\ell$ based on $\mathbf{o}^L$ and $\mathbf{y}$.
		\item[\textbf{Backward:}]
		\State Calculate $\frac{\partial \ell}{\partial \mathbf{o}^L}$;
		\For {$i=L ,L-1,\cdots,1$}
		\State Calculates $\frac{\partial\mathbf{o}^{i}}{\partial \mathbf{o}^{i-1}}$, $\frac{\partial \mathbf{o}^i}{\partial \mathbf{W}^i}$, and $\frac{\partial \mathbf{o}^i}{\partial V_{th}^i}$ by \cref{representation-all};
		\State $ \frac{\partial \ell}{\partial \mathbf{W}^i} = \frac{\partial \ell}{\partial \mathbf{o}^i}  \frac{\partial \mathbf{o}^i}{\partial \mathbf{W}^i}$;
		\State $ \frac{\partial \ell}{\partial V_{th}^i} = \frac{\partial \ell}{\partial \mathbf{o}^i}  \frac{\partial \mathbf{o}^i}{\partial V_{th}^i}$;
		\If {$i\ne 1$}
		\State $\frac{\partial \ell}{\partial \mathbf{o}^{i-1}} = \frac{\partial \ell}{\partial \mathbf{o}^{i}}  \frac{\partial \mathbf{o}^{i}}{\partial \mathbf{o}^{i-1}}$;
		\EndIf
		\State Update $\mathbf{W}^i, V_{th}^i$ based on $\frac{\partial \ell}{\partial \mathbf{W}^i}$, $\frac{\partial \ell}{\partial V_{th}^i}$.
		\EndFor
		
	\end{algorithmic}
	
\end{algorithm} 


\section{Implementation Details}
\label{train_set}
\subsection{Dataset Description and Preprocessing}

\paragraph{CIFAR-10 and CIFAR-100} The CIFAR-10 dataset~\cite{krizhevsky2009learning} contains 60,000 32$\times$32 color images in 10 different classes, which can be separated into 50,000 training samples and 10,000 testing samples. We apply data normalization to ensure that input images have zero mean and unit variance. We apply random cropping and horizontal flipping for data augmentation. The CIFAR-100 dataset~\cite{krizhevsky2009learning} is similar to CIFAR-10 except that there are 100 classes of objects. We use the same data preprocessing as CIFAR-10. These two datasets are licensed under MIT.

\paragraph{ImageNet} The ImageNet-1K dataset~\cite{deng2009imagenet} spans 1000 object classes and contains 1,281,167 training images, 50,000 validation images and 100,000 test images. This dataset is licensed under Custom (non-commercial). We apply data normalization to ensure zero mean and unit variance for input images. Moreover, we apply random resized cropping and horizontal flipping for data augmentation. 

\begin{table*}[h!tp]
	\caption{Network architectures for PreAct-ResNet-20, PreAct-ResNet-32, PreAct-ResNet-44, PreAct-ResNet-56, PreAct-ResNet-110.}
	\label{table:network}
	\centering
	\begin{tabular}{c|c|c|c|c}
		\hline  20-layers & 32-layers & 44-layers & 56-layers & 110-layers \\
		\hline \hline 
		\multicolumn{5}{c}{conv (3$\times$3,16)} \\
		\hline
		
		$\left(\begin{array}{c}3 \times 3,16 \\
		3 \times 3,16\end{array}\right) \times 3$ & $\left(\begin{array}{c}3 \times 3,16 \\
		3 \times 3,16\end{array}\right) \times 5$ & $\left(\begin{array}{c}3 \times 3,16 \\
		3 \times 3,16\end{array}\right) \times 7$ & $\left(\begin{array}{c}3 \times 3,16 \\
		3 \times 3,16\end{array}\right) \times 9$ & $\left(\begin{array}{c}3 \times 3,16 \\
		3 \times 3,16\end{array}\right) \times 18$ \\
		\hline
		
		$\left(\begin{array}{c}3 \times 3,32 \\
		3 \times 3,32\end{array}\right) \times 3$ & $\left(\begin{array}{c}3 \times 3,32 \\
		3 \times 3,32\end{array}\right) \times 5$ & $\left(\begin{array}{c}3 \times 3,32 \\
		3 \times 3,32\end{array}\right) \times 7$ & $\left(\begin{array}{c}3 \times 3,32 \\
		3 \times 3,32\end{array}\right) \times 9$ & $\left(\begin{array}{c}3 \times 3,32 \\
		3 \times 3,32\end{array}\right) \times 18$ \\
		\hline
		
		$\left(\begin{array}{c}3 \times 3,64 \\
		3 \times 3,64\end{array}\right) \times 3$ & $\left(\begin{array}{c}3 \times 3,64 \\
		3 \times 3,64\end{array}\right) \times 5$ & $\left(\begin{array}{c}3 \times 3,64 \\
		3 \times 3,64\end{array}\right) \times 7$ & $\left(\begin{array}{c}3 \times 3,64 \\
		3 \times 3,64\end{array}\right) \times 9$ & $\left(\begin{array}{c}3 \times 3,64 \\
		3 \times 3,64\end{array}\right) \times 18$ \\
		\hline
		
		\multicolumn{5}{c}{average pool, 10-d fc} \\
		\hline
	\end{tabular}
\end{table*}

\paragraph{DVS-CIFAR10} The DVS-CIFAR10 dataset~\cite{li2017cifar10} is a neuromophic dataset converted from CIFAR-10 using an event-based sensor. It contains 10,000 event-based images with resolution 128$\times$128 pixels. The images are in 10 classes, with 1000 examples in each class. The dataset is licensed under CC BY 4.0. Since each spike train contains more than one million events, we split the events into 20 slices and integrate the events in each slice into one frame. More details about the transformation could be found in \cite{fang2021incorporating}. To conduct training and testing, we separate the whole data into 9000 training images and 1000 test images. Both the event-to-frame integrating and data separation are handled with the SpikingJelly \cite{SpikingJelly} framework.
We also reduce the spatial resolution from 128$\times$128 to 48$\times$48 and apply random cropping for data augmentation.

\subsection{Batch Normalization}
Batch Normalization (BN) \cite{ioffe2015batch} is a widely used technique in the deep learning community to stabilize signal propagation and accelerate training. In this paper, BN is adopted in the network architectures. However, since the input data for SNNs have an additional time dimension when compared to input image data for ANNs, we need to make the BN components suitable for SNNs. 

In this paper, we combine the time dimension and the batch dimension into one and then conduct BN. In detail, consider a batch of temporal data $\mathbf{x}\in\mathbb{R}^{B\times N}$ with batch size $B$ and temporal dimension $N$ such that $\mathbf{x}=(\mathbf{x}^{(1)},\cdots,\mathbf{x}^{(B)})$, and $\mathbf{x}^{(i)}\in\mathbb{R}^N$ for $i=1,2,\cdots,N$.
Then define ${\mu}$ and ${\sigma}^{2}$ to be the mean and variance of the reshaped data $({x}^{(1)}[1],\cdots,{x}^{(1)}[N],\cdots,{x}^{(B)}[1],\cdots,{x}^{(B)}[N])$. With the defined ${\mu}$ and ${\sigma}^{2}$, BN transforms the original data $\mathbf{x}^{(i)}$ to $\hat{\mathbf{x}}^{(i)}$ as
\begin{equation} \label{bn1}
\hat{\mathbf{x}}^{(i)}=\gamma \frac{\mathbf{x}^{(i)}-\mu}{\sqrt{\sigma^2+\epsilon}}+\beta,
\end{equation}
where $\gamma$ and $\beta$ are learnable parameters, and $\epsilon$ is a small positive number to guarantee valid division.

\subsection{Network Architectures}
We use the pre-activation ResNet-18 \cite{he2016identity} network architecture to conduct experiments on CIFAR-10, CIFAR-100, and ImageNet. To make the network architecture implementable on neuromorphic chips, we add spiking neurons after pooling operations and the last fully connected classifier. Furthermore, we replace all max pooling with average pooling. To stabilize the (weighted) firing rates of the output layer, we also introduce an additional BN operation between the last fully connected classifier and the last spiking neuron layer. The network contains four groups of basic block \cite{he2016identity} structures, with channel sizes 64, 128, 256, and 512, respectively.

To test the effectiveness of the proposed DSR method with deeper networks structures, we conduct experiments with the pre-activation ResNet with 20, 32, 44, 56, 110 layers, whose architectures are shown in \cref{table:network}. We also add additional spiking neuron layers and BN layers like what we do for the PreAct-ResNet-18 structure.

We use the VGG-11 \cite{simonyan2014very} network architecture to conduct experiments on DVS-CIFAR10. To enhance generalization capacity of the network, we further add dropout \cite{srivastava2014dropout} layers after the spiking neurons, and we set the probability of zeroing elements to be $0.1$.  We only keep one fully connected layer to reduce the number of neurons.

\subsection{Training Hyperparameters}
First, we consider hyperparameters about the IF model. We set the initial threshold for each layer $V_{th}^i=6$ and restrict it to be no less than $0.01$ during training. We set $\alpha$ in \cref{modify-reset} to be $0.5$.

Then we consider hyperparameters about the LIF model. We fix the time constant $\tau^i$ for the $i^{\text{th}}$ layer to be $1$ for each $i$. The setting for initial $V_{th}^i$, lower bound for $V_{th}^i$, $\Delta t$, and $\alpha$ change with the number of time steps, as shown in \cref{table:lif-setting}.
\begin{table}[H]
	\caption{Hyperparameters for the LIF model given different number of time steps. ``LB for $V_{th}^i$'' means lower bound for $V_{th}^i$.}
	\label{table:lif-setting}
	\centering
	\begin{tabular}{c|cccc}
		\hline Time Steps & initial $V_{th}^i$ & LB for $V_{th}^i$ & $\Delta t$ & $\alpha$  \\
		\hline 20 & 0.3 & 0.0005 & 0.05 & 0.3 \\
		15 & 0.3 & 0.0005 & 0.05 & 0.4 \\
		10 & 0.3 & 0.0005 & 0.05 & 0.4 \\
		5 & 0.6 & 0.001 & 0.1 & 0.5 \\
		\hline
	\end{tabular}
\end{table}

Next, we consider hyperparameters about the optimization. We use cosine annealing \cite{loshchilov2016sgdr} as the learning rate schedule for all datasets. Other hyperparameters can be found in \cref{table:parameter}. Also note that we change the initial learning rate from $0.1$ to $0.05$ when using 5 time steps.

\begin{table}[H]
	\caption{Hyperparameters about Optimization for training CIFAR-10, CIFAR-100,  ImageNet, and DVS-CIFAR10.}
	\label{table:parameter}
	\hspace{-0.8em}
	\begin{tabular}{c|cccc}
		\hline Dataset & Optimizer & Epoch & $lr$   & Batchsize\\
		\hline CIFAR-10 & SGD \cite{rumelhart1986learning} & 200 & $0.1$  & 128  \\
		CIFAR-100 & SGD \cite{rumelhart1986learning} & 200 & $0.1$  & 128  \\
		ImageNet & Adam \cite{kingma2014adam} & 90 & $0.001$ & 144  \\
		DVS-CIFAR10 & SGD \cite{rumelhart1986learning} & 300 & $0.05$ & 128  \\
		\hline
	\end{tabular}
\end{table}

\section{Firing Sparsity}
\begin{figure}[h]
	\centering
	\begin{subfigure}{0.8\linewidth}
		\includegraphics[width=1\columnwidth]{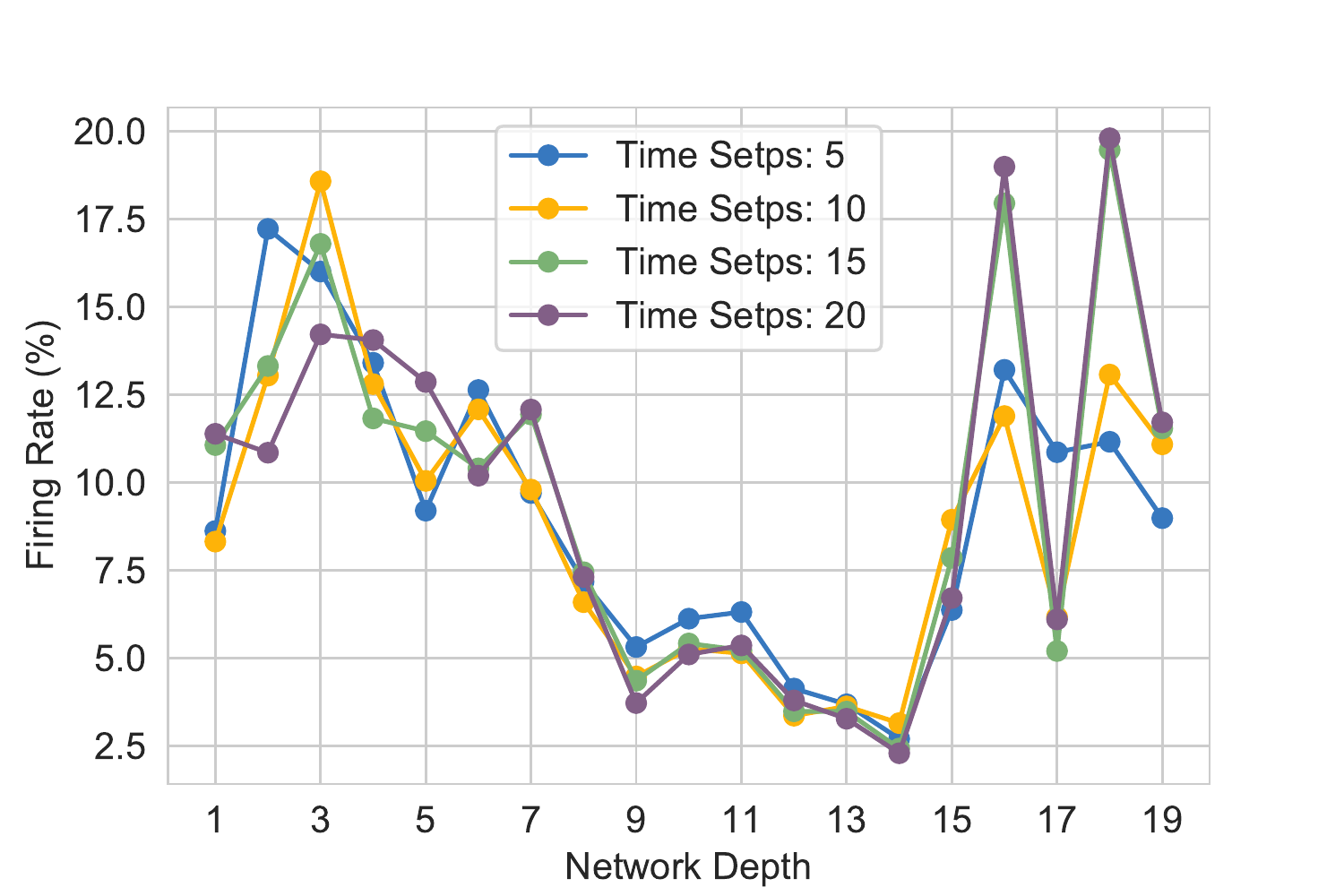} 
		\caption{Firing rates for the IF model.}
		\label{fig:firing-if}
	\end{subfigure}
	
	\vspace{1em}
	
	\begin{subfigure}{0.8\linewidth}
		\includegraphics[width=1\columnwidth]{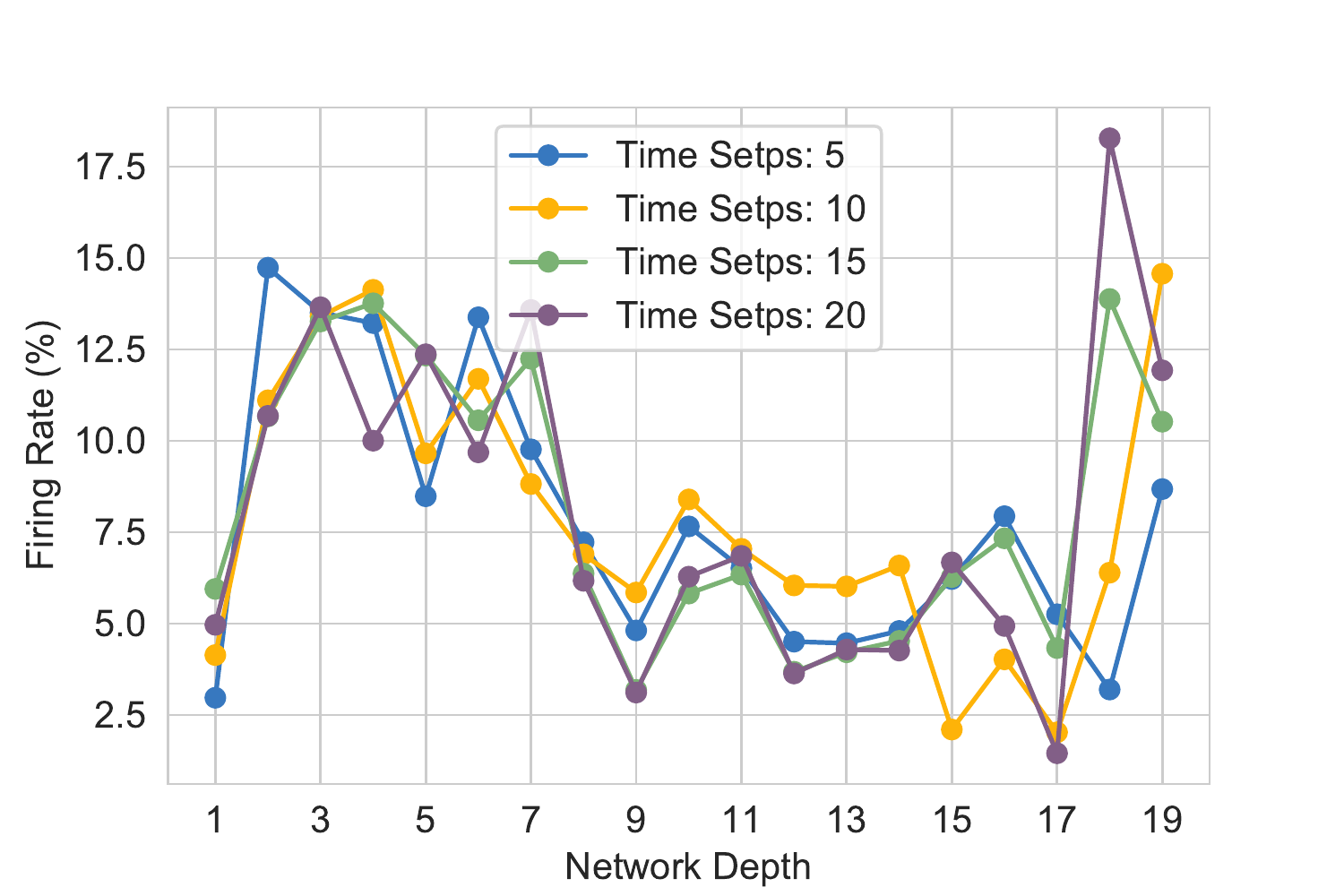} 
		\caption{Firing rates for the LIF model.}
		\label{fig:firing-lif}
	\end{subfigure}
	
	\caption{Firing rates of the trained SNNs by the proposed DSR method. The SNNs are trained on CIFAR-10 with the PreAct-ResNet-18 structure and different number of time steps. Firing rates of different spiking neuron layers are shown.}
	\label{fig:firing}
\end{figure}

To achieve low energy consumption on neuromorphic hardware, the number of spikes generated by an SNN should be small. Then the firing rate is an important quantity to measure the energy efficiency of SNNs. We calculate the average firing rates of the trained SNNs on CIFAR-10, as shown in \cref{fig:firing}. The results show that the firing rates of all layers are below 20\%, and many layers have firing rates of no more than 5\%. Regardless of the layer of neurons, the total firing rates are between 7.5\% and 9.5\% for different number of time steps and both neuron models. Since the firing rate does not increase as the number of time steps decreases, the proposed method can achieve satisfactory performance with both low latency and high firing sparsity.

\section{Weight Quantization}
In our experiments, the network weights are 32-bit. However, we can also adopt low-bit weights when implementing our method on neuromorphic hardware, by combining existing quantization algorithms. The weights in neuromorphic hardware are generally 8-bit. So we simply quantize the weights of our trained SNNs to 8 bits and even 4 bits using the straight-through estimation method, and the results on CIFAR-10 are shown in \cref{table:quantization}.
\begin{table}[h]
	\caption{Performances on CIFAR-10 with network weights of different precisions. The PreAct-ResNet-18 architecture with 20 time steps is used.}
	\label{table:quantization}
	\centering
	\begin{tabular}{c|ccc}
		\hline Neural Model & 32 bits & 8 bits & 4 bits \\
		\hline 
		IF & 95.38\% & 95.45\% & 95.31\% \\
		LIF & 95.63\%& 95.65\% & 95.39\% \\
		\hline
	\end{tabular}
	\vspace{-5pt}
\end{table}\\

\end{document}